\documentclass{article}

\usepackage{times}
\usepackage{graphicx} 
\usepackage{subfigure} 

\usepackage{natbib}

\usepackage{algorithm}
\usepackage{algorithmic}

\usepackage{booktabs}

\usepackage{tikz}
\usepackage{pgfplots}
\pgfplotsset{compat=1.3}
\usepackage{filecontents}
\usetikzlibrary{matrix}
\usetikzlibrary{calc}
\usetikzlibrary{arrows}
\usetikzlibrary{patterns}
\usetikzlibrary{plotmarks}
\usetikzlibrary{intersections} 
\usetikzlibrary{decorations.pathmorphing}
\usetikzlibrary{decorations.pathreplacing}
\usetikzlibrary{decorations.markings,arrows} 
\usetikzlibrary{decorations.shapes}
\usetikzlibrary{fit}
\usetikzlibrary{backgrounds}
\usetikzlibrary{trees}

\usepackage{hyperref}

\usepackage[accepted]{icml2016}

\icmltitlerunning{Efficiency Evaluation of Character-level RNN Training Schedules}

\begin{document} 

\twocolumn[
\icmltitle{Efficiency Evaluation of Character-level RNN Training Schedules}

\icmlauthor{Cedric De Boom, Sam Leroux, Steven Bohez, Pieter Simoens,}{}
\icmlauthor{Thomas Demeester, Bart Dhoedt}{firstname.lastname@ugent.be}
\icmladdress{Ghent University -- iMinds, Department of Information Technology, Technologiepark 15, 9052 Zwijnaarde, Belgium}

\icmlkeywords{efficiency, RNN, neural networks, deep learning, machine learning, text, characters}

\vskip 0.2in
]

\begin{abstract} 
We present four training and prediction schedules from the same character-level recurrent neural network.
The efficiency of these schedules is tested in terms of model effectiveness as a function of training time and amount of training data seen.
We show that the choice of training and prediction schedule potentially has a considerable impact on the prediction effectiveness for a given training budget.
\end{abstract} 

\section{Introduction}
\label{sec:introduction}
Recurrent neural networks (RNNs) are able to take a data sequence of arbitrary length as input, map it to a hidden state, and then use this hidden state to make a prediction.
This prediction can be a single value, but it can also be a entire new sequence.
In this paper we will study a `character-level RNN', i.e.~a language model that learns to predict the next character of a text given the history of characters seen.
At every time step a new character is fed into the RNN, which then transforms its hidden state to be able to predict the next character.
Such a type of RNN is also used in applications such as speech to text translation \cite{Graves:2014vz} and video frame tagging \cite{Pigou:2015vg}.
In the next section will illustrate how such a model can be learned using different training and prediction schedules, and how this affects the model efficiency.

\section{Experimental setup}
\label{sec:setup}
In all experiments from this section we will use the following neural network architecture:

Input (65 dimensions) -- LSTM layer (50 dimensions, input/output/forget gate non-linearity: sigmoid, cell non-linearity: tanh) -- Dense layer (65 dimensions, softmax non-linearity).

The number of input and output dimensions is 65, since there are 65 different characters in our text corpus, and we encode every character in a one-hot representation.
We choose to use a Long Short-Term Memory (LSTM) layer above a simple recurrent connection as this is currently state of the art in many text mining tasks \cite{Greff:2015wv}.


\begin{figure*}
\centering
\includegraphics[width=\linewidth]{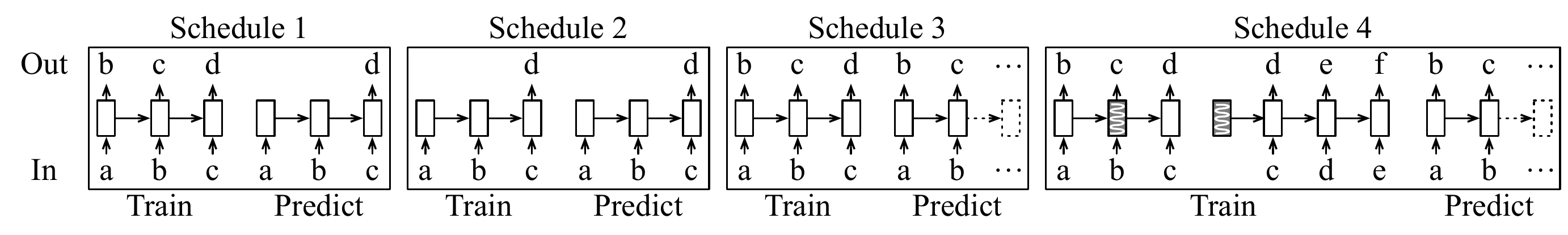}
\caption{Four different schedules of training and predicting from a character-level RNN.}
\label{fig:schedules}
\end{figure*}

\begin{figure*}
\centering
\includegraphics[trim = 0.85cm 0.2cm 1.5cm 1.2cm, clip, width=0.4\linewidth]{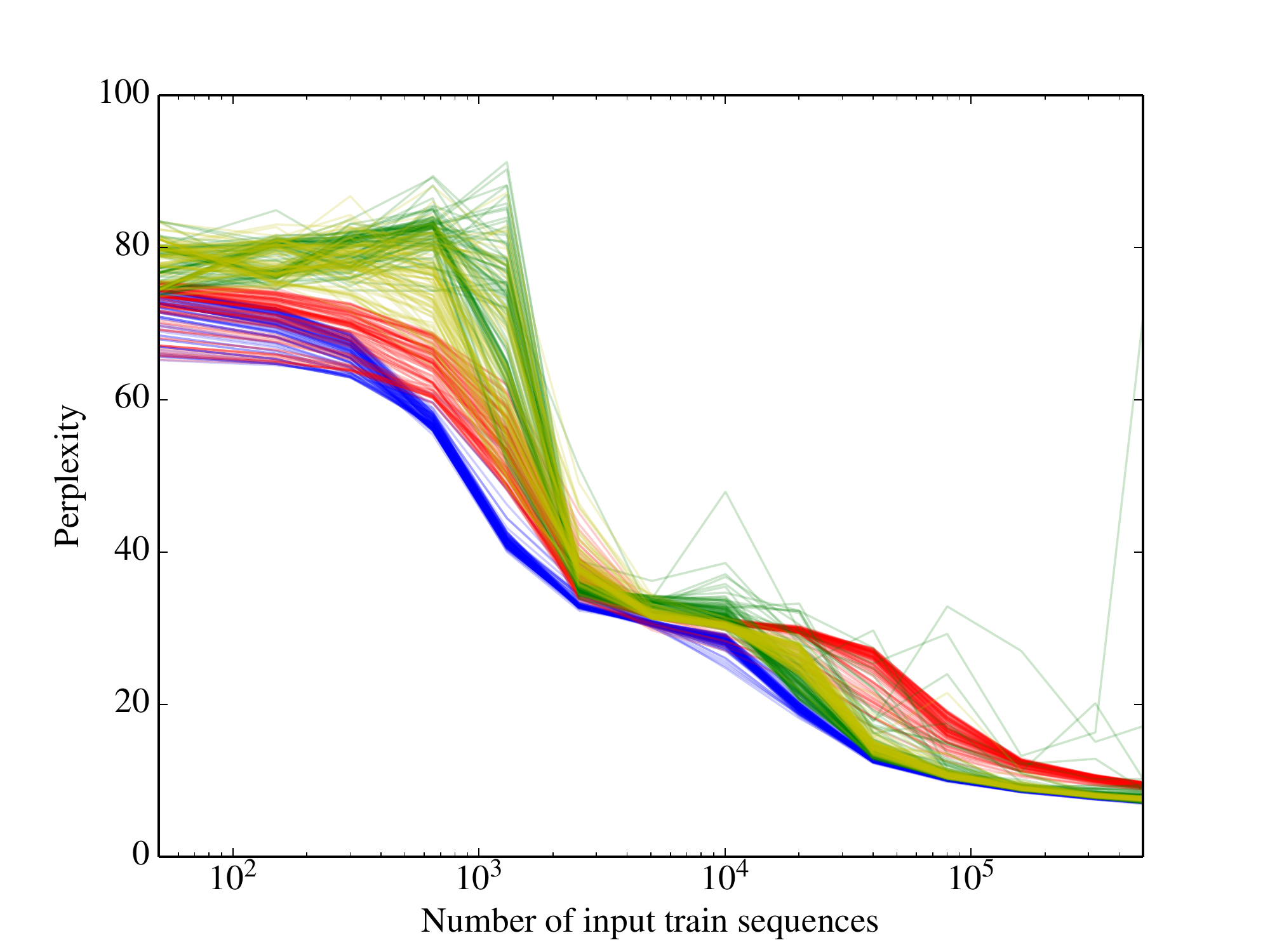}
\includegraphics[trim = 0.85cm 0.2cm 1.5cm 1.2cm, clip, width=0.4\linewidth]{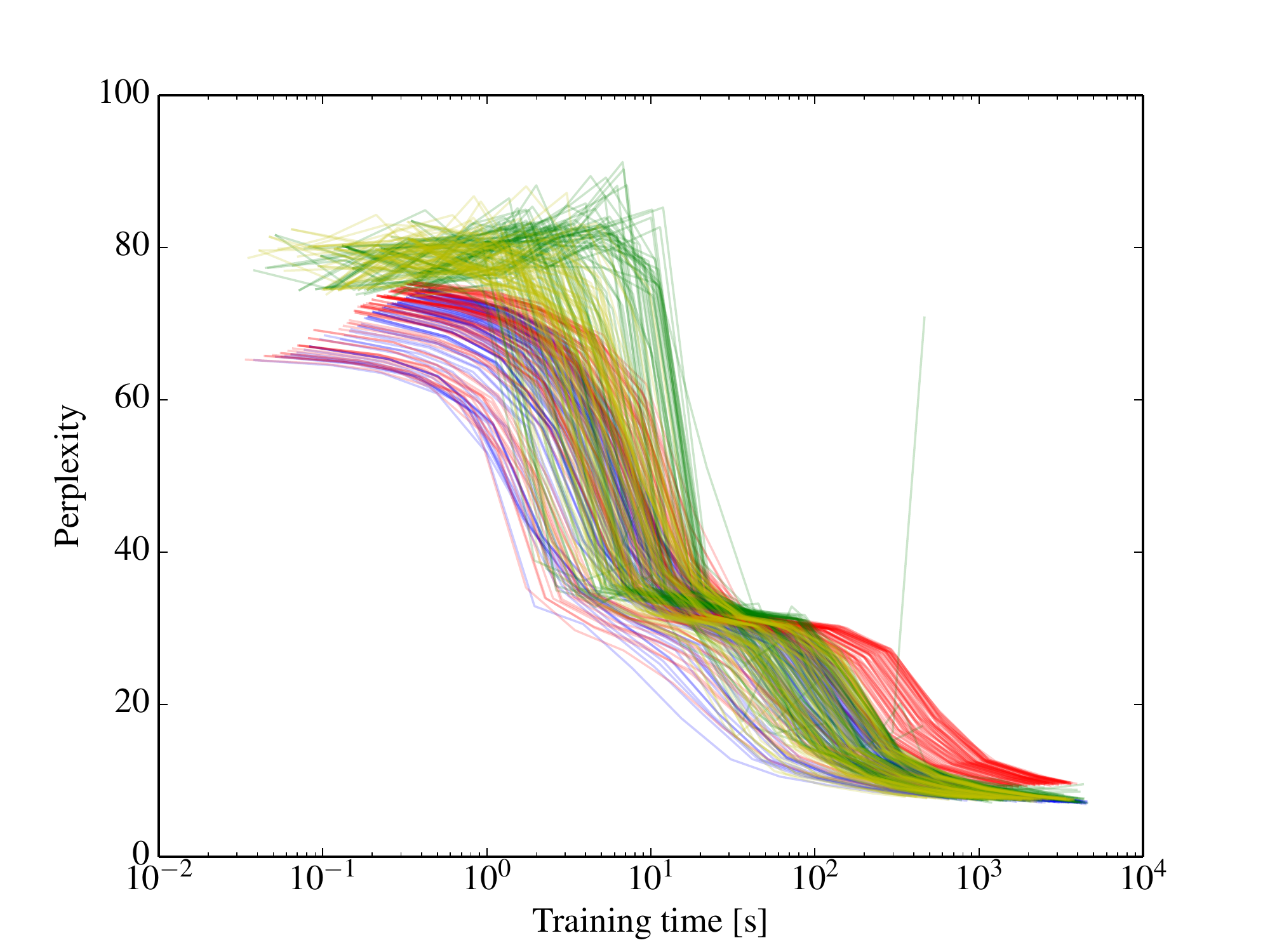}
\caption{Perplexity as a function of number of input train sequences and elapsed training time. Blue: schedule 1; red: schedule 2; green: schedule 3; yellow: schedule 4.}
\label{fig:batch}
\end{figure*}

To train this neural network we use Truncated Backprogratation Through Time (TBPTT).
Every $k_1$ time steps the gradients are backpropagated for $k_2$ time steps in the past (we use the same parameter notation as in \cite{Sutskever:2013wo}).
To allow for temporal dependencies in the input sequence between every gradient update, we can additionally ensure that $k_1 \leq k_2$.
The choice of these parameters can be a trade-off between computation time and model effectiveness.
To calculate the gradients in the TBPTT, we use the categorical cross entropy as loss function.

Apart from choosing these parameters, there is a choice of how to implement the training and prediction procedures, which can influence both model effectiveness and training speed.
Below we list four different schedules for the character-level RNN.
A visualization of each of the schedules is shown in Figure \ref{fig:schedules}.

Schedule 1 --
In this schedule we take sequences of fixed length $k_2$ as input for training, and for every input character in such a sequence we predict the next character.
This leads to $k_2$ losses that we backpropagate through the network.
The hidden and cell states in the LSTM layer are reset to their initial values for every input sequence.
These initial states are learned through backpropagation.
At test time, we take an entire sequence of length $k_2$ as input, and we ask to predict the single next character.

Schedule 2 --
This schedule is similar to the first one, but instead of making a prediction for every input character at train time, we now only predict the final next character.
We therefore only have a single loss that is backpropagated.
At test time, the procedure is the same as in schedule 1.

Schedule 3 --
Here, the training procedure is the same as in the first schedule.
At test time we start predicting using the learned initial hidden state.
After that the subsequent character is predicted for every input character, after which the hidden state is updated to be used in the next prediction.

Schedule 4 --
At train time, the initial hidden state for a new input sequence is reused from the previous input sequence.
More specifically, as initial hidden state for the input character at time $t$ we use the hidden state produced by input character at time $t-1$ from the previous input sequence.
The prediction procedure in this final schedule is the same as in schedule 3.

To evaluate the predictive capacity of a character-level RNN, we use the perplexity measure, traditionally used to measure the effectiveness of language models.
We will evaluate the four schedules explained above in terms of perplexity vs.~training time and training input sequences: the most efficient implementations should reach a lower perplexity faster than the other ones.
Every implementation is run a 100 times with random settings for the $k_2$ and $k_1$ parameters: $5 \leq k_1 \leq k_2 \leq 50$.
This is all done on the same hardware, and for a total of 500,000 input train sequences.
We employ Adam gradient updates \cite{Kingma:2015ku} with a batch size of 50 across all experiments, and we use the Lasagne implementation framework (\url{github.com/Lasagne/Lasagne}).
Our dataset consists of excerpts from Shakespearian plays; the train set has around 1,100,000 characters, and the test set around 11,000.

It is clear from the figures that all settings for schedules 1 and 2 converge smoothly towards an optimum, but schedule 1 is more efficient.
Schedules 3 and 4 have noisy behaviour in the beginning of the training phase.
After that all settings for schedule 4 seem to converge to the same optimum, but some parameter settings for schedule 3 continue to behave very noisily.
Schedule 1, however, seems to be performing best and most consistently.
We observe that the settings for which the lowest perplexity is reached the fastest, all have a small $k_2 < 10$.
Since $k_1 \leq k_2$, this means that frequent model updates over short input sequences are preferred.

Experimental code and data can be found on: \url{https://github.com/cedricdeboom/CharRNN}.

\section{Conclusion}
We tested the efficiency of multiple training and prediction schedules of a character-level recurrent neural network, in terms of model effectiveness as a function of training time and the number of training input sequences.
We observed that the choice of a particular schedule can considerably impact the efficiency of the model.
It also turns out that training over short input sequences and with frequent model updates is most efficient.
Further research is however required to verify if these conclusions hold for more complex models and other datasets.

\section*{Acknowledgements} 
We thank Pontus Stenetorp for his useful feedback on drafts of this paper.
We also acknowledge Nvidia for its generous hardware support.
Cedric De Boom is funded by a PhD~grant of the Flanders Research Foundation (FWO). Steven Bohez is funded by a PhD~grant of the Agency for Innovation by Science and Technology in Flanders (IWT). 

\bibliographystyle{icml2016}

\end{document}